%% file: root.tex
\documentclass[letterpaper, 10 pt, conference]{ieeeconf}  

\IEEEoverridecommandlockouts                              

\usepackage{graphics} 
\usepackage{graphicx}
\usepackage{amsmath} 
\usepackage{amssymb}  
\usepackage{xcolor}
\usepackage{subfigure}
\usepackage{siunitx}
\usepackage{threeparttable}
\makeatletter
\let\NAT@parse\undefined
\makeatother
\usepackage[colorlinks,
            linkcolor=blue,
            anchorcolor=blue,
            citecolor=blue]{hyperref}
\usepackage{algorithm}
\usepackage{algorithmicx}
\usepackage{tabularx,booktabs}
\usepackage{multirow}

\newcommand{\ie}{\textit{i}.\textit{e}.}

\title{\LARGE \bf
EdgeCalib: Multi-Frame Weighted Edge Features for Automatic Targetless LiDAR-Camera Calibration
}

\author{Xingchen Li$^{1}$, Yifan Duan$^{1}$, Beibei Wang$^{2}$, Haojie Ren$^{1}$, Guoliang You$^{1}$, Yu Sheng$^{1}$, \\ Jianmin Ji$^{1}$, Yanyong Zhang$^{1}$
\thanks{$^1$ School of Computer Science and Technology, University of Science and Technology of China (USTC), Hefei 230026, China}
\thanks{$^2$ Institute of Artificial Intelligence, Hefei Comprehensive National Science Center, Hefei, Anhui, China}
\thanks{ $^\dag$ Corresponding author. {\tt\small yanyongz@ustc.edu.cn}}}

\begin{document}

\maketitle
\thispagestyle{empty}
\pagestyle{empty}

\begin{abstract}
In multimodal perception systems, achieving precise extrinsic calibration between LiDAR and camera is of critical importance. 
Previous calibration methods often required specific targets or manual adjustments, making them both labor-intensive and costly. Online calibration methods based on features have been proposed, but these methods encounter challenges such as imprecise feature extraction, unreliable cross-modality associations, and high scene-specific requirements.
To address this, we introduce an edge-based approach for automatic online calibration of LiDAR and cameras in real-world scenarios. The edge features, which are prevalent in various environments, are aligned in both images and point clouds to determine the extrinsic parameters. Specifically, stable and robust image edge features are extracted using a SAM-based method and the edge features extracted from the point cloud are weighted through a multi-frame weighting strategy for feature filtering. Finally, accurate extrinsic parameters are optimized based on edge correspondence constraints. 
We conducted evaluations on both the KITTI dataset and our dataset. The results show a state-of-the-art rotation accuracy of 0.086° and a translation accuracy of 0.977 cm, outperforming existing edge-based calibration methods in both precision and robustness.

\end{abstract}

\input{1_introduction}

\input{2_related}

\input{3_methods}
\input{4_experiment}

\input{5_conclusion}

\bibliographystyle{IEEEtran}
\bibliography{IEEEabrv, refs}

\end{document}

%% file: 1_introduction.tex
\section{Introduction}
\label{sec:intro}
In autonomous intelligent systems, such as self-driving vehicles, multimodal perception (e.g., LiDAR and camera) is typically required for robust and accurate perception of the environment~\cite{zhu2022vpfnet,li2022mathsf, graeter2018limo}. LiDAR sensors offer precise but sparse three-dimensional geometric information, while cameras provide dense color information but lack accurate depth data. The fusion of these two modalities can yield complementary benefits. However, effectively calibrating the extrinsic parameters between these sensors presents a significant challenge.

Existing offline sensor calibration methods often rely on explicit targets, such as chessboards~\cite{zhang2004extrinsic}. 
However, these methods are not only cumbersome and costly but also lack the ability for real-time correction of parameter shifts due to prolonged operations and load variations.
Therefore, various calibration methods have been proposed to find feature correspondences across sensors in an online and real-time fashion, and many methods accomplish this by leveraging manually extracted features, such as geometric or semantic cues~\cite{levinson2013automatic, wang2020soic}. 
Nonetheless, these techniques, which heavily depend on precise feature extraction, might be compromised by random factors, such as sensor noise. 
Moreover, these methods rely heavily on the extraction of rich features, making them less effective in sparse environments.

To address this feature extraction challenge, large-scale deep learning models offer a promising solution. The recently released the Segment Anything Model (SAM)~\cite{kirillov2023segment} is a visual foundation model for image segmentation, which demonstrated impressive zero-shot capabilities, benefiting from large amounts of training data, thus making it well-suited for various vision tasks, including edge detection. In this work, we employ SAM to facilitate a precise edge contour extraction from individual image frames. 

\begin{figure}[t]
    \centering
    \includegraphics[width=\linewidth]{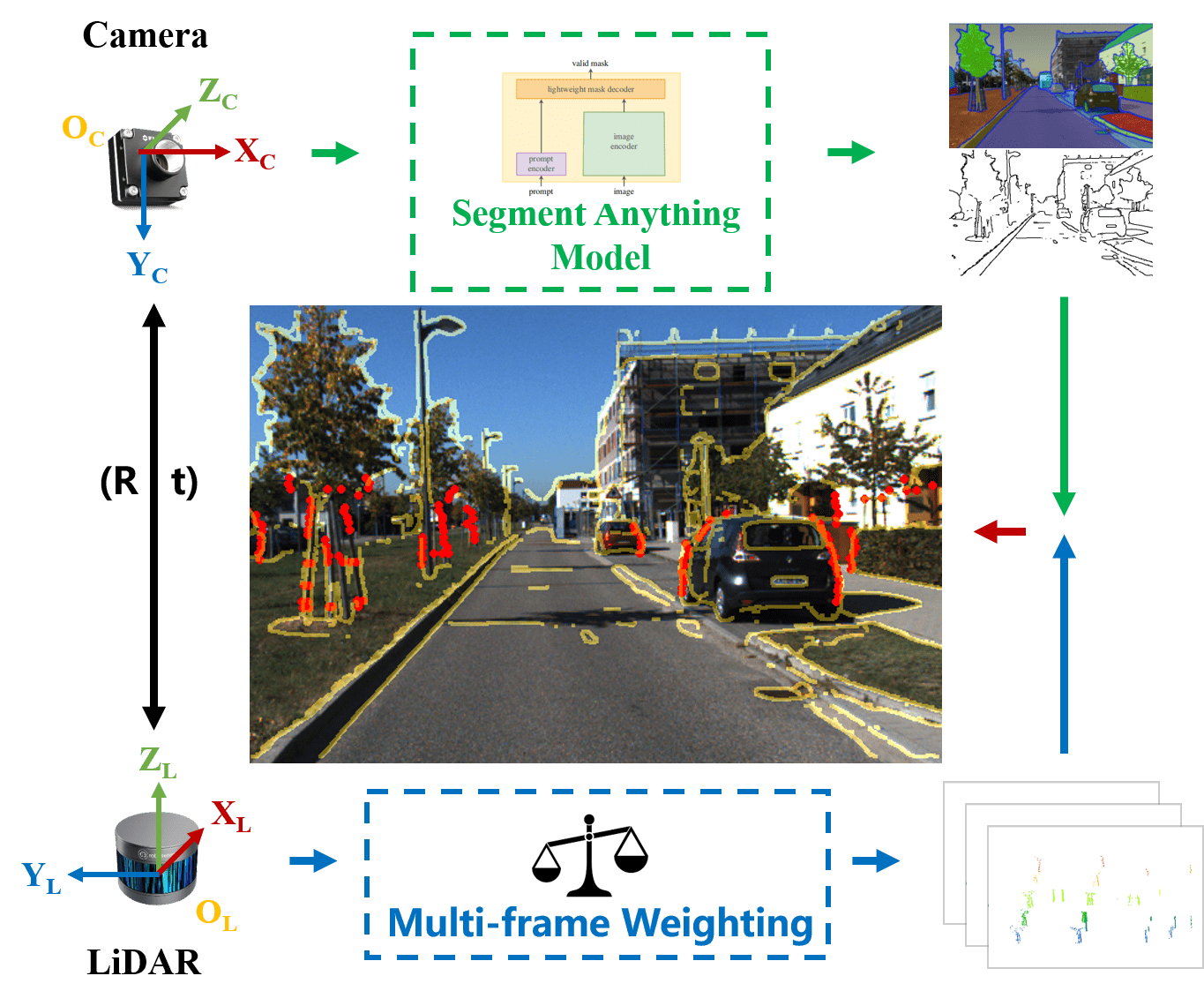}
    \caption{The core principle of our approach is to exploit edge consistency for calibration. Specifically, we apply the SAM to process camera data, while employing a multi-frame weighting strategy for LiDAR data processing.}
    \label{fig:intro}
    \vspace{-1.5em}
\end{figure}

Another identified limitation in existing calibration methods is their exclusive focus on single-frame point cloud-image pairs, lacking sufficient exploration of inter-frame feature distributions.~\cite{levinson2013automatic, zhang2021line, yuan2021pixel}. To explore the consistency of point edges across multiple frames, we further extend our method to incorporate multi-frame features via single-frame feature extraction and matching. We explore position consistency and projection consistency of edge features across sequential frames to further optimize feature selection. Fig.~\ref{fig:intro} illustrates the underlying idea of our method. 

The main contributions of our work can be listed as follows.
\begin{enumerate}
    \item We introduce an automatic target-less method for LiDAR-camera extrinsic calibration, leveraging edge consistency between point clouds and images. By utilizing SAM, along with an adaptive edge filtering strategy, our approach significantly enhances the accuracy of edge contour extraction. 
    \item We introduce a multi-frame feature weighting strategy, enhancing the stability and robustness of edge feature extraction. This significantly improves the calibration accuracy of extrinsic parameters.
    \item Our method achieves state-of-the-art performance with a rotation error of $0.086\si{\degree}$ and a translation error of $0.977\, \si{\cm}$. Furthermore, we validate the generalizability and versatility of our approach through experiments on the publicly available KITTI dataset, as well as on our collected data.
\end{enumerate}

The rest of this article is organized as follows: Section~\ref{sec:related} summarizes related works. Section~\ref{sec:methods} introduces the methods proposed in this article. Section~\ref{sec:exper} evaluates the accuracy of the proposed methods based on the dataset. Section~\ref{sec:conclusion} summarizes our research and prospects for future work.


%% file: 2_related.tex
\section{related works}
\label{sec:related}
Depending on the requirement for explicit calibration targets, strategies for extrinsic alignment between a LiDAR sensor and a camera can be bifurcated into two distinct categories, \ie\ target-based and target-less methods.

\subsection{Target-based methods}
Target-based methods usually use explicit and regular objects as indicators for the correspondence between point clouds and images. The application of a calibration target for LiDAR-camera calibration was first introduced by~\cite{zhang2004extrinsic}. They proposed to use a checkerboard from multiple views to calibrate a 2D LiDAR sensor and a camera. In this method, the extrinsic parameters were estimated by solving a nonlinear least-square iterative minimization problem. However, they require manual annotation in LiDAR data processing.

Researchers have made efforts to avoid human intervention in calibration, thereby enhancing the overall intelligence of the process.
\cite{geiger2012automatic} presented a single-shot, multi-object automatic extrinsic calibration method. Specifically, their method required finding several checkerboards at different sites in one image of the scene, instead of taking multiple shots of a single checkerboard and changing its location.

The target-based calibration is restricted to laboratory settings due to the necessity for specific targets. Moreover, complex scene configurations and human involvement frequently impede the effectiveness of these methods to handle extrinsic parameter variations on moving vehicles.

\begin{figure*}[t]
    \centering
    \includegraphics[width=0.9\linewidth]{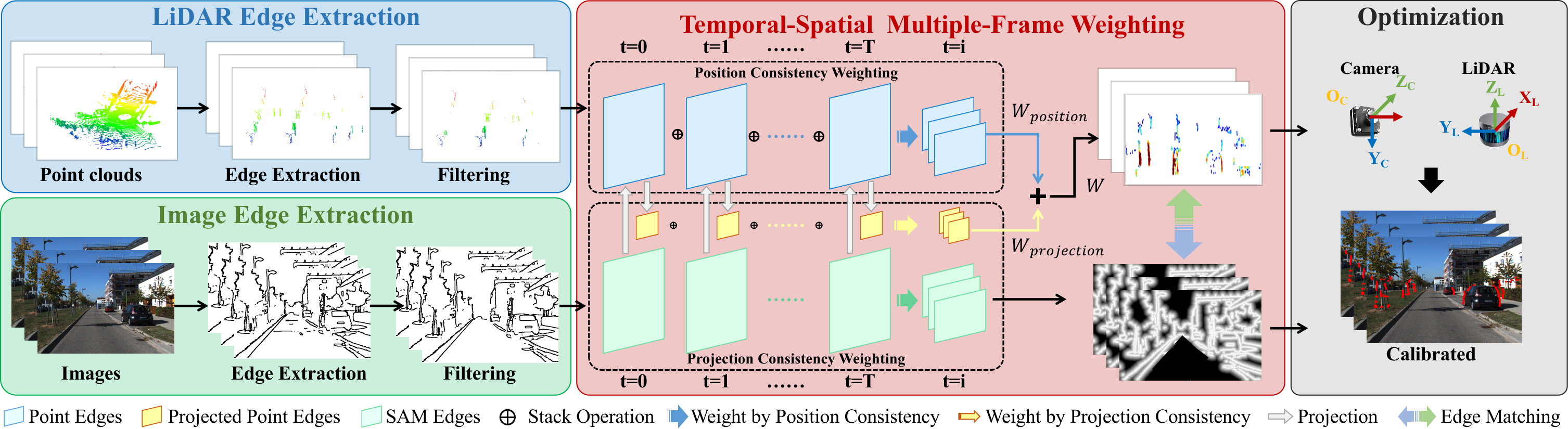}
    \caption{An overview of the proposed method.}
    \label{fig:pipline}
    \vspace{-1.0em}
\end{figure*}

\subsection{Target-less methods}
In real-world applications, such as autonomous driving, online and real-time estimation of extrinsic parameters becomes necessary. In such scenarios, target-less calibration methods are employed to retrieve extrinsic parameters without the need for specific targets, by leveraging valuable information from the surrounding environment automatically. These approaches promises to eliminate the need for designated calibration targets and are able to minimize manual labor requirements.

Target-less methods can be further categorized into four types: information-theory-based, feature-based, ego-motion-based, and learning-based~\cite{li2022survey}. 
Information-theory-based methods~\cite{pandey2012automatic, pandey2015automatic, wang2012automatic} estimate the extrinsic parameters by maximizing the similarity transformation between the LiDAR sensor and the camera, as measured by various information metrics.

Ego-motion-based methods exploit the motion of sensors mounted on the moving vehicle to estimate the extrinsic parameters. Z. Taylor et al.~\cite{taylor2015motion} extended the traditional hand-eye calibration method to address the calibration problem between LiDAR sensors and cameras.  
Learning-based methods do not require the artificial definition of features and can learn useful information directly through neural networks. 
In the works~\cite{schneider2017regnet, iyer2018calibnet, lv2021lccnet}, network models are employed to process input from both camera images and LiDAR point clouds, subsequently directly outputting the extrinsic parameters.

In feature-based approaches, features are extracted and paired between LiDAR data and camera images. Levinson et al.~\cite{levinson2013automatic} proposed an online calibration method based on the observation that the depth discontinuities identified by the LiDAR sensor are likely to coincide with image edges. 
Later, \cite{zhang2021line} applied feature filtration to refine the feature extraction, and utilized an adaptive optimization method to search for the optimal direction and obtain the extrinsic parameters. The work by \cite{liao2023se} employs semantic edges, but the semantic edge detection network in their approach faces generalization problems.

%% file: 3_methods.tex
\section{METHODOLOGY}
\label{sec:methods}
The core issue in the extrinsic calibration of the LiDAR sensor and the camera is to effectively establish correspondence between data from different modalities. In this study, edge features are selected as the fundamental elements for accurate extrinsic calibration between the LiDAR and camera coordinate systems. These edge features, which include objects like trees, street lights, and walls, are commonly found in both indoor and outdoor environments, and they have the potential to demonstrate consistent correspondences between point clouds and images. 

The challenge in the extrinsic calibration between the LiDAR sensor and the camera lies in precisely determining the transformation matrix that correlates them. In the scope of this paper, we formalize this problem as the identification of the transformation matrix $\mathbf{T}_L^C \in S E(3)$, which is composed of a rotation matrix and a translation vector.
We make assumptions that the intrinsic parameters of the camera and LiDAR are already well-calibrated, and that the data from LiDAR and camera sensors are captured at the same moment.

Fig.~\ref{fig:pipline} offers an overview of the proposed approach. 
First, edge features are extracted and initially filtered from images and point clouds. Then, a multi-frame weighting strategy is utilized to further optimize the edges. Finally, point cloud edge features are projected onto pixel frames, during which the score of current parameter is calculated and optimized.

\subsection{Image edges extraction}
\label{sec:method_image edge}
In online calibration methods based on edge features, image edges are typically extracted using edge detectors such as the Canny Edge Detector~\cite{canny1986computational}. However, traditional algorithms like Canny primarily rely on local image gradients and orientations for edge detection, lacking an overall understanding of image semantics. As a result, the extracted edges often include a lot of texture and noise, which can make them relatively cluttered.

In contrast to conventional approaches that utilize geometric edge detectors for edge detection, we exploit the extensive capabilities of the large-scale visual model, \ie, SAM, for accomplishing edge detection tasks. Serving as a deep learning foundational model, SAM exhibits robust zero-shot performance across a diverse range of segmentation tasks. Specifically, we employ SAM by prompting it with a 16×16 grid of foreground points to produce predicted masks, which are subsequently refined using methods such as non-maximal suppression and Sobel filtering to generate the edge maps efficiently. 

While edges extracted using SAM excel at capturing object contours, we observe the presence of some ``weak edges'' arising from low-confidence semantic segmentation. 
These weak edges often represent textures or noise, such as roads or grasslands, and generally lack a corresponding relationship with LiDAR-generated edges. Therefore, they need to be filtered out through a specific strategy.

Therefore, after obtaining the edge maps generated by SAM, we employ an adaptive edge filtering strategy based on semantic information. For each object mask generated by SAM, we extract its contour and calculate the standard intensity of the edge pixels along it. This standard intensity serves as an adaptive threshold for retaining high-confidence edges within each object's segmented region. Consequently, the adaptive approach yields a more nuanced edge map, prioritizing edges that correspond to object boundaries over texture features.

\begin{figure}[t]
\begin{center}
	\subfigure[Original image]{
	\begin{minipage}{\linewidth}
        \centerline{\includegraphics[width=\textwidth]{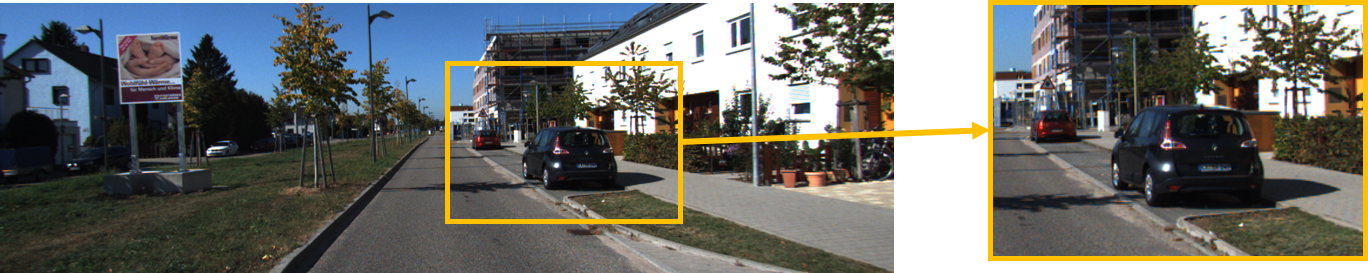}}
	\end{minipage}
        }
        \subfigure[Canny edges]{
	\begin{minipage}{\linewidth}
        \centerline{\includegraphics[width=\textwidth]{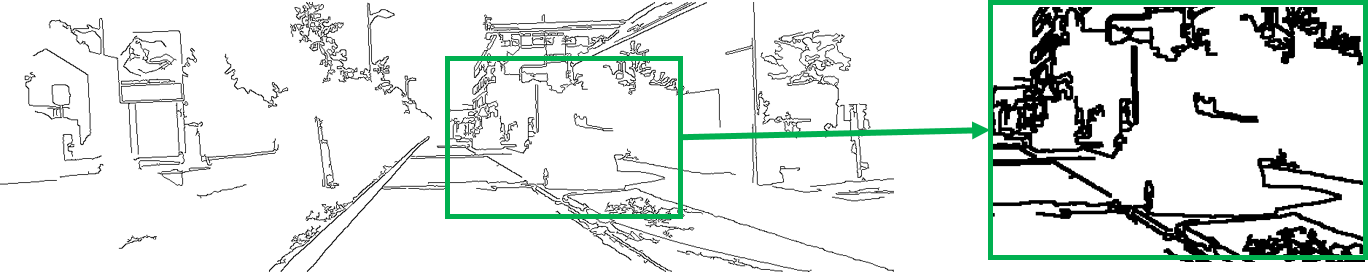}}
	\end{minipage}
	}
        \subfigure[SAM edges]{
        \begin{minipage}{\linewidth}
        \centerline{\includegraphics[width=\textwidth]{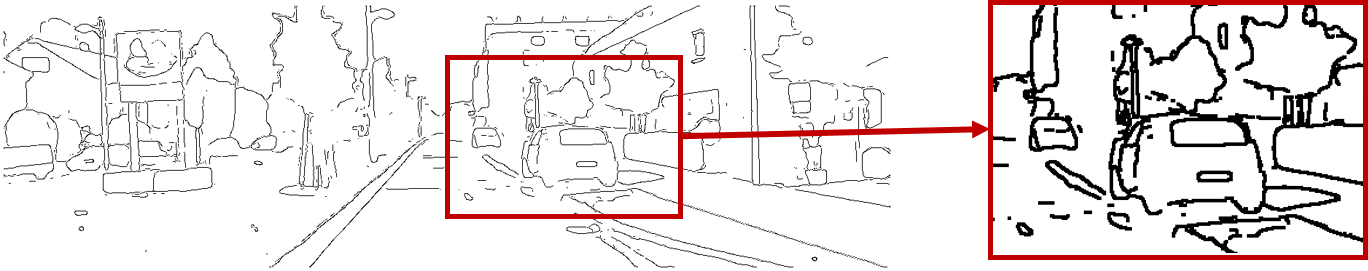}}
	\end{minipage}
	}
	\caption{A comparison of edge features extracted by Canny and SAM. It can be observed that SAM edges outperform those obtained by the Canny method, particularly in capturing object contours.}
\label{fig:image_edges}
\end{center}
\vspace{-2.5em}
\end{figure}

We refer to the edges extracted by the SAM model as ``SAM edges." Examples of an original image, Canny edge features, and SAM edge features are shown in Fig. \ref{fig:image_edges}. 
It is evident that, due to the semantic information that assists in understanding the image's content, SAM edge features enable more accurate extraction of object boundaries. This results in fewer incorrectly extracted texture edges compared to traditional Canny edge features.

The correspondence between a single edge point and an image edge pixel is unstable and may have many outliers. Inspired by the work of Levinson~\cite{levinson2013automatic}, we construct an edge attraction field map, denoted as $G_i, i=1, \ldots, n$, generated via a distance transform applied to the image edge map. In accordance with distance transform properties, each pixel in $G_i$ encodes its $L2$ distance to the nearest image edge. Consequently, the negative gradient of this attraction field map inherently indicates edge correspondence, guiding the projected points towards improved edge alignment.

\subsection{LiDAR edges extraction}
In LiDAR data processing, distance discontinuities are utilized to extract edge features within point clouds. Given the lower point cloud density along the vertical direction of the scan line, the analysis primarily focuses on the horizontal depth difference between consecutive scans, thereby enabling the detection of edges within the structure of point clouds.

For a given LiDAR point, the distance between the current point and its adjacent counterpart is examined. If a sudden depth variation is detected, the current point is designated as an edge point. Concurrently, points with shallower depths at the laser depth discontinuity are selected, as they remain visible from the camera's perspective, unlike deeper points that may not be. This mitigates potential issues caused by disparate fields of view between the sensors.

Currently, the edge features extracted from point clouds are often unstructured and disorganized. To refine this, a point cloud clustering algorithm is applied to eliminate edge features with insufficient adjacent points. This filtering stage leads to a more coherent and well-organized representation of point cloud edge features, thereby facilitating enhanced optimization results in subsequent processes. 
To simplify writing, we refer to the edge features extracted from point cloud as ``point edges".

\subsection{Multi-frame weighting strategy}
\label{sec:method_weight}
\begin{figure}[t]
\begin{center}
	\subfigure[Position Consistency]{
	\begin{minipage}{0.46\linewidth}
 
	\centerline{\includegraphics[width=\textwidth]{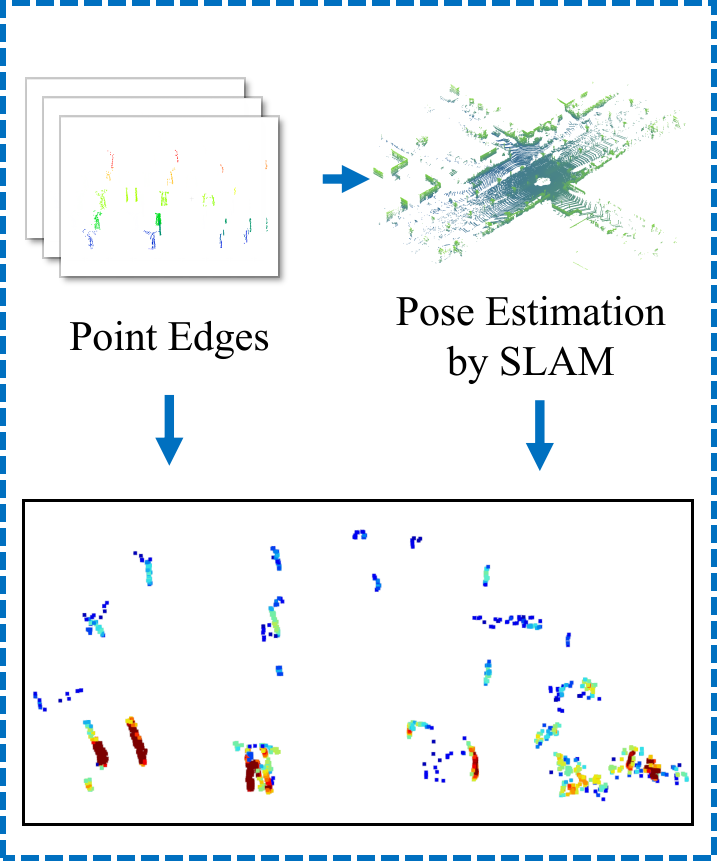}}
        \label{fig:position}
	\end{minipage}
        }
        \subfigure[Projection  Consistency]{
	\begin{minipage}{0.46\linewidth}
 
        \centerline{\includegraphics[width=\textwidth]{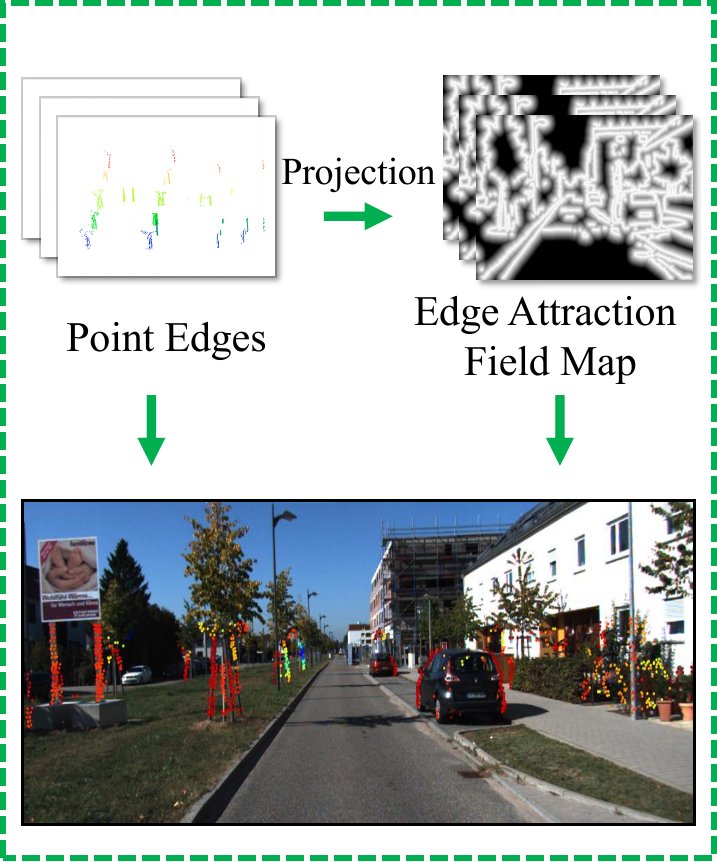}}
        \label{fig:projection}
	\end{minipage}
	}
	\caption{Illustration of the position consistency and projection consistency. Red means high weight and blue means low weight for the two types of consistency.}
\label{fig:two_consistency}
\end{center}
\vspace{-2.0em}
\end{figure}

Using SAM edges in images and edges in point clouds, we can achieve a single-frame calibration results. However, to obtain more information beneficial to calibration, we analyze the continuous data sequence from LiDAR to explore the consistency of point edges across multi-frame, which weights point edges in the optimization step. Consequently, we propose two attributes, i.e., position consistency and projection consistency, of point edges to further improve the calibration accuracy, as shown in Fig.\ref{fig:two_consistency}.

First of all, it's obvious that the distribution of edges is spatially consistent under ideal conditions. That is, when the positions of the LiDAR captures data are close, e.g., two adjacent frames, the point cloud sampling from the surface of one actual edge-shaped object can be extracted as a point edge with great probability.
We use this property, named as position consistency, to reduce the impact of incorrectly extracted edge features in the subsequent optimization problem. In specific, to correlate data at different frames, we use poses of each frames to align and stack the point edges as a local edge map, denoted as $\mathcal{M}$. Then, we define a weight, denoted as $W_{position}$ to quantitatively describe this position consistency as follow:

\begin{small}
\begin{equation}
\begin{aligned}
    W_{position}(p_i^t) &= \sum_{p\in\mathcal{M}}[d(p, p^t_i) < r],\\
\end{aligned}
\label{eq:w_position}
\end{equation}
\end{small}
where $p_i^t$ means the i-th point in t-th frame of point edges, $d(p, p^t_i)$ means the distance between the point $p$ and $p_i^t$, $[P]$ is Iverson bracket~\cite{iverson1962programming} which is equal to 1 when $P$ is true and $r$ represents the radius, which is set in advance. Eq.~\ref{eq:w_position} computes the local density at $p^t_i$ in the map, which represents the local temporal-spatial distribution of point edges and can be viewed as position consistency.

On the other hand, the above-mentioned position consistency cannot directly bring positive contributions to the calibration, for the reason that the accuracy of calibration is only related to the correspondence between image edges and point edges but not the position consistency which aims to improve the quality of point edges. Therefore, we additionally design projection consistency. Specifically, we first project each point edges in the sequence on the edge attraction field map introduced in Sec.~\ref{sec:method_image edge} and assign the grayscale values  corresponding to each point. Subsequently, similar to $W_{position}$, a quantitative attribute, denoted as $W_{projection}$, describing the projection consistency is defined as follows:
\begin{small}
\begin{equation}
    W_{projection}(p_i^t) = \sum_{p\in\mathcal{M}} G(p)[d(p, p^t_i) < r],
\label{eq:w_position}
\end{equation}
\end{small}
where $G(p)$ means the gray value obtained by $p$ during the above projection process. Please note that this is a simplified notation and the specific calculation process will be introduced in Sec.~\ref{sec:method_optimization}. The high value of $W_{projection}$ means that the points in the neighborhood of $p^t_i$ usually have a corresponding relationship with the edges of the image.

Finally, the weight of edge points in each frame is calculated as
\begin{small}
\begin{equation}
    W(p_i^t) = \alpha W_{position}(p_i^t)+\beta W_{projection}(p_i^t).
\label{eq:w_weight}
\end{equation}
\end{small}

\begin{figure*}[t]
\begin{center}
	\subfigure[]{
	\begin{minipage}{0.45\linewidth}
 
	\centerline{\includegraphics[width=\textwidth]{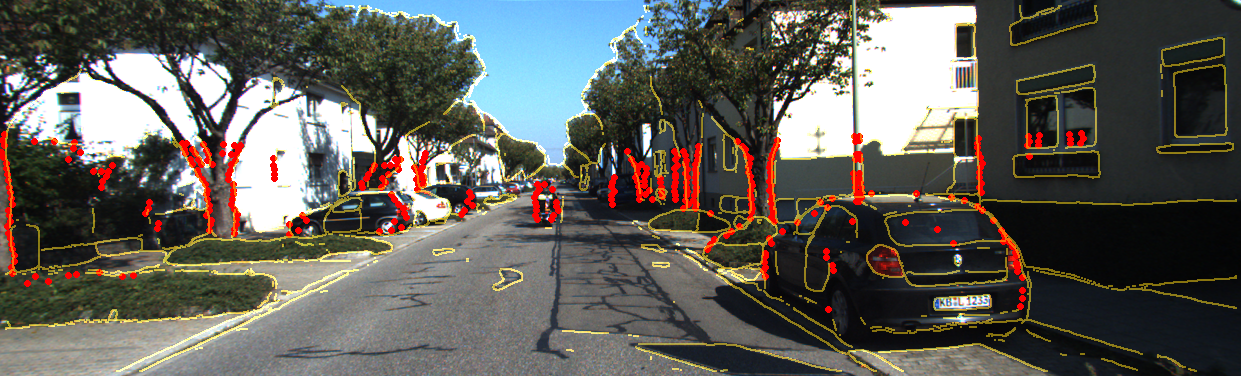}}
        \label{fig:proj_1}
	\end{minipage}
        }
        \subfigure[]{
	\begin{minipage}{0.45\linewidth}
 
        \centerline{\includegraphics[width=\textwidth]{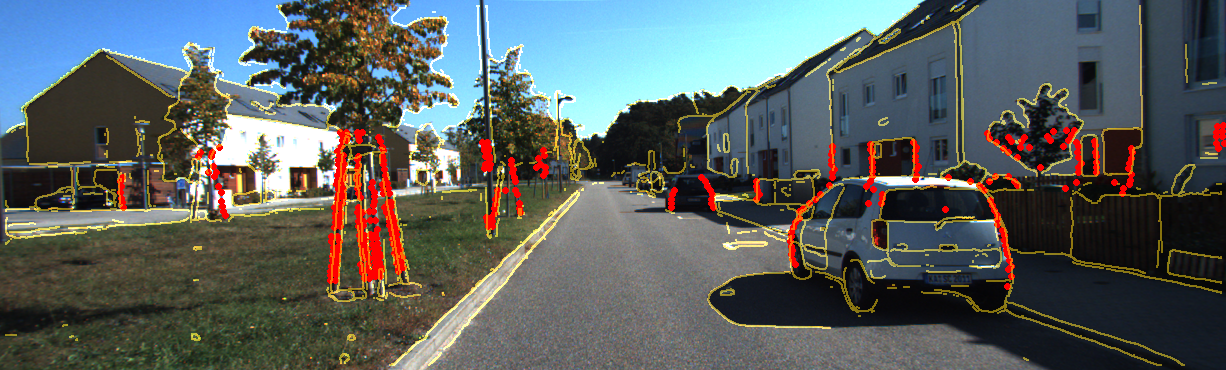}}
        \label{fig:proj_2}
	\end{minipage}
	}
	\caption{The qualitative calibration results. The red point cloud edges are accurately projected onto the gold image edges.}
\label{fig:visualization}
\end{center}
\vspace{-2.0em}
\end{figure*}


\subsection{Multi-frame edge alignment optimization}
\label{sec:method_optimization}
The extracted LiDAR edges need to be matched to their corresponding edges in the image.
In this step, we calibrate the extrinsic parameters by minimizing the reprojection error of aligned edges. 

We first define the projection function, that project points in LiDAR frames onto images. Let the coordinate of a point in the LiDAR coordinate system be $  p_l = [ x_L\ y_L\ z_L \ 1]^\top $ and the coordinate of a pixel on the image plane be $ P_C = [ u \ v \ 1]^\top$. $\mathbf{T_L^C}$ denotes the extrinsic parameters between the LiDAR and camera coordinates, $K$ denotes the intrinsic parameters of the camera, and $s$ denotes the depth scale factor. Then,

\begin{small} 
\begin{equation}
s \left[\begin{matrix} u \\ v \\ 1 \end{matrix}\right]= K\left( \mathbf{T}_L^C \left[\begin{matrix} x_L \\ y_L \\ z_L \\ 1\end{matrix}\right] \right)_{1: 3}.
\end{equation}
\end{small}

For the point clouds and images, we use the notation $P^t$ for point clouds and $I_{i j}^t$ for images, where $i j$ specifies the pixel values at the image coordinates onto which point $p_l$ is projected at frame $t$. 
We aim to optimize the extrinsic parameters by projecting each point in the point cloud onto the corresponding image frame.

The score for the current extrinsic parameter is calculated by summing the grayscale values of projected points, thus formulating the objective function, \ie, 

\begin{small} 
\begin{equation}
\operatorname{S}_t(\mathbf{T}_L^C)=\sum_{t=n-w}^n \sum_{p_l}^{p_l \in F_e^t} W(p_l)G_i^t \left[ K \left(\mathbf{T}_L^C p_l \right)_{1:3} \right],
\end{equation}
\end{small}
where each LiDAR point $p_l$ iterates edge features $F_e$ and the $W(p_l)$ is defined in Sec.~\ref{sec:method_weight}.  
In addition, $w$ is the size of the sliding window. Score $S_t$ of frame $t$ is computed by considering the previous $w$ frames.

The calibration problem can be viewed as a function of the thermal extrinsic $T_L^C$, and the correct calibration $T^*$ can be estimated by

\begin{small}
\begin{equation}
T^* = \arg \min _\mathbf{T_L^C}\left\|\operatorname{S}_t(\mathbf{T}_L^C)\right\|^2.
\end{equation}
\end{small}

We can transform the optimization problem to an unconstrained form with Lie algebra. Let $\mathbf{T_L^C}=\exp \left(\xi^{\wedge}\right), \xi \in$ $\mathfrak{s e}(3)$. Then, the error term of a edge point $p_l \in F_e$ is

\begin{small}
\begin{equation}
e(\xi)=G_i\left(\frac{1}{s} K \exp \left(\xi^{\wedge}\right) p_l\right) .
\end{equation}
\end{small}
The derivative of $e(\xi)$ can be found by applying a small left disturbance to $\xi$, \ie,

\begin{small} 
\begin{equation}
\begin{aligned}
& \lim _{\delta \xi \rightarrow 0} e(\xi \oplus \delta \xi) \\
& \quad=\lim _{\delta \xi \rightarrow 0} G_i\left(\frac{1}{s} K \exp \left(\xi^{\wedge}\right) p_l+\frac{1}{s} K \delta \xi^{\wedge} \exp \left(\xi^{\wedge}\right) p_l\right) .
\end{aligned}
\end{equation}
\end{small}

Let $p{,}=\delta \xi^{\wedge} \exp \left(\xi^{\wedge}\right) p_l$ and $u=\left(1 / s\right) K p_{,}$, then

\begin{small} 
\begin{equation}
\begin{aligned}
& \lim _{\delta \xi \rightarrow 0} e(\xi \oplus \delta \xi) 
=\lim _{\delta \xi \rightarrow 0} e(\xi)+\frac{\partial G_i}{\partial u} \frac{\partial u}{\partial p^{\prime}} \frac{\partial p^{\prime}}{\partial \delta \xi} \delta \xi .
\end{aligned}
\end{equation}
\end{small}

So, the Jacobian matrix of the error $e(\xi)$ is

\begin{small}
\begin{equation}
J_{\xi}=\frac{\partial G_i}{\partial u} \frac{\partial u}{\partial p^{\prime}} \frac{\partial p^{\prime}}{\partial \delta \xi}.
\end{equation}
\end{small}

where $\left(\partial G_i / \partial u\right)$ is the image gradient of $G_i$ at $u$, and the rest part is the standard 3-D to 2-D projection model.


With the Jacobian matrix, we can easily adopt the Levenberg-Marquardt method to solve this nonlinear optimization problem.

%% file: 4_experiment.tex
\section{EXPERIMENTS}
\label{sec:exper}
\subsection{Dataset preparation}

\begin{figure}[t]
    \centering
    \includegraphics[width=0.8\linewidth]{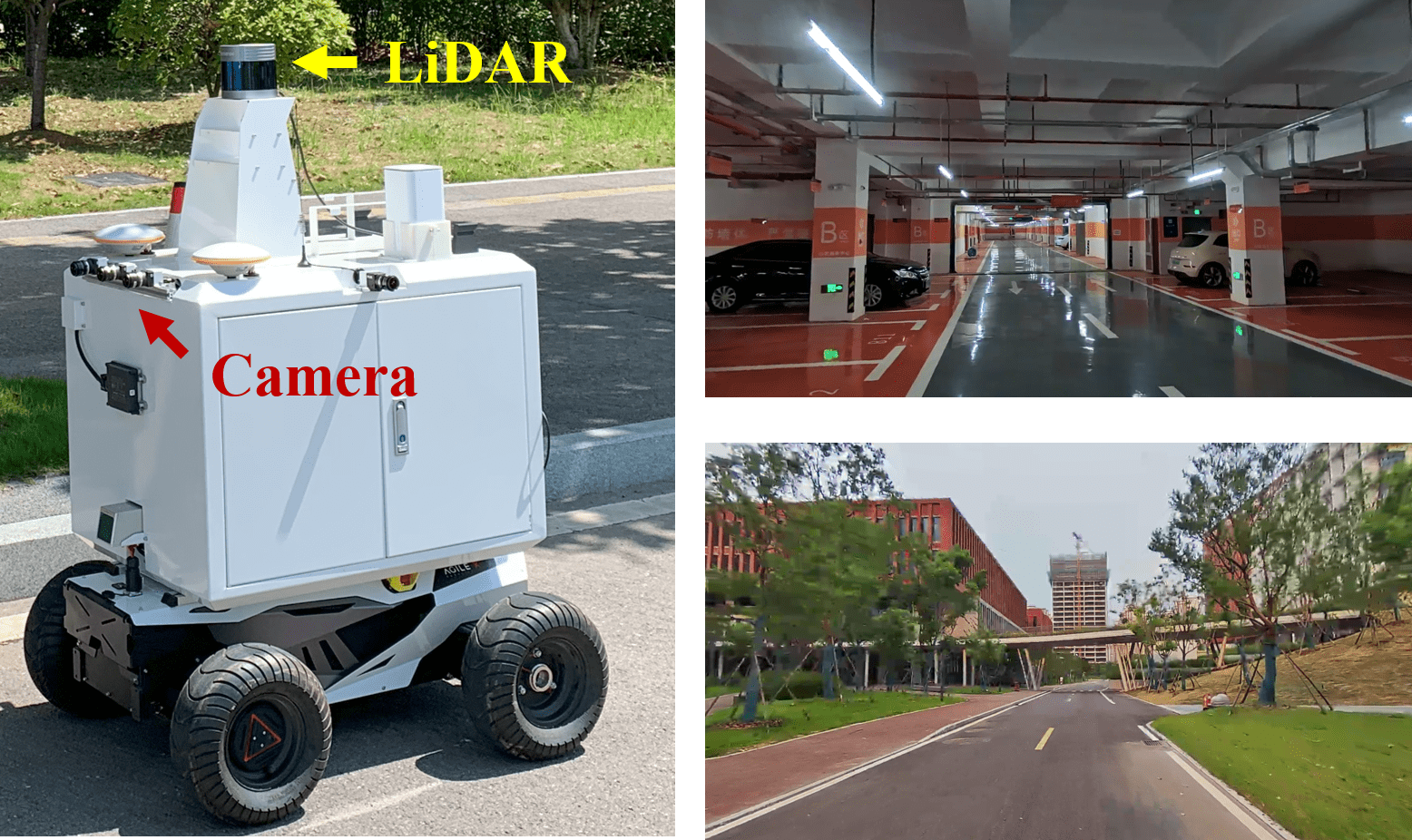}
    \caption{On the left is our autonomous vehicle equipped with a LiDAR sensor and camera, while on the right are the datasets we collected, featuring both indoor and outdoor environments.}
    \label{fig:ourdata}
\end{figure}

We perform experiments on two datasets. The first dataset is derived from the KITTI Odometry Benchmark~\cite{geiger2013vision}, which includes a Velodyne HDL-64E LiDAR and a high-resolution color camera with a scanning frequency of 10 Hz. The data used for the experiment is synchronized and rectified, and the ground truth for extrinsic parameters can be obtained from the provided calibration files.
The second one is collected from our autonomous vehicle, equipped with a RoboSense RS-LiDAR-32 LiDAR sensor and a color camera. The dataset, synchronized at a rate of 10 Hz and rectified, consists of 5986 frames containing both images and point clouds. Our acquisition device is shown in the Fig.~\ref{fig:ourdata}.

\begin{table}[tbp]
    \caption{Comparison results with other edge-based calibration methods}
    \label{tab:comparison}
    \centering
    \begin{tabularx}{\linewidth}{c *{4}{X}}
        \toprule
        Methods & Mean(\si{\degree}) $\downarrow$ & Roll(\si{\degree}) $\downarrow$ & Pitch(\si{\degree}) $\downarrow$ & Yaw(\si{\degree}) $\downarrow$ \\
        \midrule
        Levinson~\cite{levinson2013automatic}  & $1.043$ & $0.991$ & $1.067$ & $1.037$ \\
        Zhang\cite{zhang2021line} & $0.517$ & $0.493$ & $0.452$ & $0.487$ \\
        SE-Calib~\cite{liao2023se} & $0.215$ & $0.180$ & $0.223$ & $0.206$ \\
        Ours & $\mathbf{0.086}$ & $\mathbf{0.124}$ & $\mathbf{0.036}$ & $\mathbf{0.097}$ \\
        \bottomrule
    \end{tabularx}
\end{table}

\begin{figure}[t]
    \centering
    \includegraphics[width=0.9\linewidth]{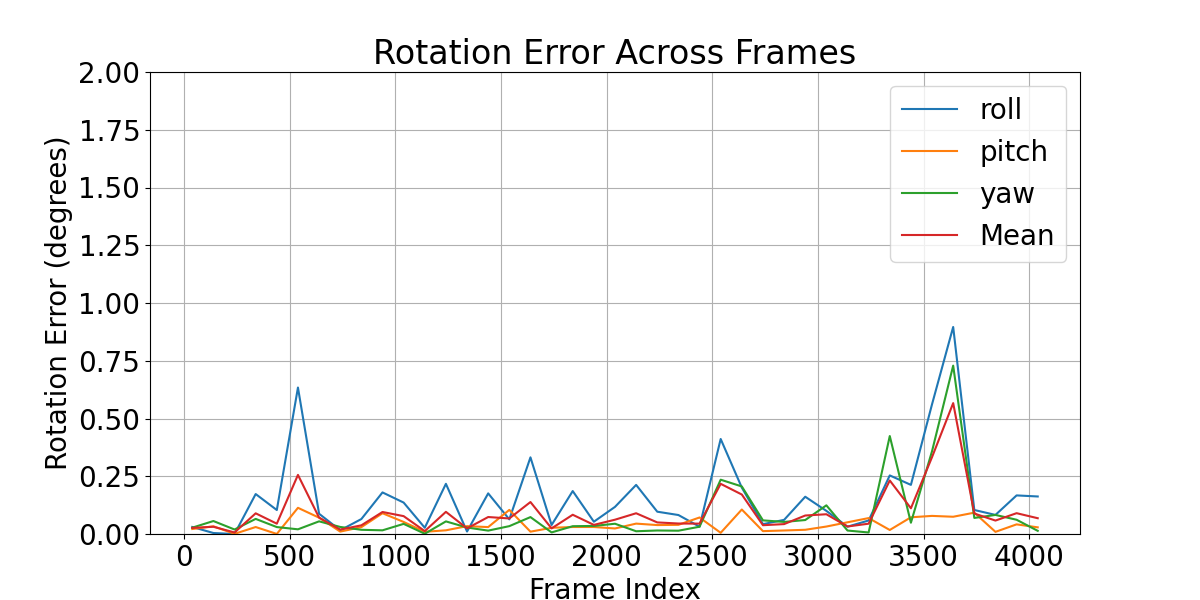}
    \caption{The distribution of rotation errors across the entire dataset.}
    \label{fig:plot}
    \vspace{-1.5em}
\end{figure}

\subsection{Quantitative results}

To evaluate the feasibility of our proposed method, we conduct evaluations on two distinct sequences (sequences 0 and 7) from the KITTI odometry dataset, which include different scenes. 

In the experiment, we initially add a $2 \si{\degree}$ rotation bias on X, Y, and Z axes and a $10\,\si{\cm}$ transformation bias to the ground truth parameters, where the sign of perturbation is randomized. Each experiment utilizes an aggregate of 100 frames, approximately 10 seconds of data. During the experiment, we compare the calibration error with the ground truth. 

The results are as follows: Our proposed method achieves a mean translation error of $\mathbf{0.977}\, \si{\cm}$ (x, y, z: $1.168\, \si{\cm}$, $1.256\, \si{\cm}$, $0.507\, \si{\cm}$) and a mean rotation error of $\mathbf{0.086\si{\degree}}$ (roll, pitch, yaw: $0.124 \si{\degree}$, $0.036 \si{\degree}$, $0.097 \si{\degree}$). 

We conducted a comparative analysis of our proposed algorithm against three similar methods, namely, Levinson~\cite{levinson2013automatic}, Zhang~\cite{zhang2021line}, and SE-Calib~\cite{liao2023se}. These methods utilize edge features, straight line features, and semantic edge features, respectively. 
Since these methods only report results on rotation parameters, we compared the accuracy of rotation parameters. As illustrated in Tab.~\ref{tab:comparison}, our method exhibits superior performance over these state-of-the-art algorithms when tested on the KITTI dataset. Compared to the three edge-based methods mentioned above, our results have achieved improvements of approximately $\mathbf{91.75\%}$, $\mathbf{83.38\%}$, and $\mathbf{59.07\%}$, respectively. The distribution of rotation errors is illustrated in Fig.~\ref{fig:plot}, showing that the errors remain relatively stable in most instances.

To further demonstrate the performance of our method, we also tested initial parameter offsets that were much larger than the settings of previous edge-based methods. Specifically, under rotation offsets of $5 \si{\degree}$ and $10 \si{\degree}$, our method achieves a rotation error of $0.122 \si{\degree}$ and $0.412 \si{\degree}$, respectively. Likewise, under translation offsets of $50 \si{\cm}$ and $100 \si{\cm}$, our method delivers a translation error of $1.065 \si{\cm}$ and $1.733 \si{\cm}$.


\subsection{Qualitative results}
To visualize the calibration results, we project the edge point cloud onto the image plane in ``red'' using the computed extrinsic parameters. Concurrently, SAM edges appear as ``golden'' lines on the same image, as illustrated in Fig.~\ref{fig:visualization}. It can be observed that both sets of edges, representing object contours, exhibit good alignment.

\begin{table}[tp]
    \caption{Ablation experiments on SAM edge and multi-frame weighting strategies.}
    \label{tab:ablation}
    \centering
    \begin{tabularx}{\linewidth}{c *{4}{X} *{4}{X}}
        \toprule
        \multirow{2}[2]{*}{Setting} & \multicolumn{4}{c}{Translation(\si{\cm}) $\downarrow$} & \multicolumn{4}{c}{Rotation(\si{\degree}) $\downarrow$} \\
        \cmidrule(lr){2-5} \cmidrule(lr){6-9}
         & Mean & X & Y & Z & Mean & Roll & Pitch & Yaw \\ 
        \midrule
        Ours & $\mathbf{0.977}$ & $\mathbf{1.168}$ & $\mathbf{1.256}$ & $\mathbf{0.507}$ & $\mathbf{0.086}$ & $\mathbf{0.124}$ & $\mathbf{0.036}$ & $\mathbf{0.097}$\\
        w/o SAM & $8.364$ & $9.941$ & $9.877$ & $5.275$ & $1.489$ & $2.663$ & $0.325$ & $1.480$\\
        w/o weighting & $2.461$ & $2.335$ & $3.309$ & $1.739$  & $0.222$ & $0.466$ & $0.058$ & $0.144$ \\
        \bottomrule
    \end{tabularx}
    \vspace{-1.0em}
\end{table}

\subsection{Ablation studies}
In this section, we compare the impact of several modules to the overall method.

\vspace{2pt}\noindent\textbf{Impact of SAM edges.}
In our image feature extraction step, we employ semantic edge features derived from SAM for calibration. This is different from most previous work that uses geometric edges \cite{levinson2013automatic, zhang2021line, yuan2021pixel}. To evaluate the significance of using SAM edges in our work, we also implement the conventional Canny edge detector for image edge feature extraction. Operating under the same test conditions and dataset, 
as presented in Table~\ref{tab:ablation}, our SAM edge-based approach demonstrates a $90.30\%$ reduction in translation error and a $94.20\%$ reduction in rotation error compared to the Canny-based approach. The result shows the greater precision and robustness of SAM edges over geometric edges.

\vspace{2pt}\noindent\textbf{Impact of multi-frame weighting.}
Leveraging SAM allows us to extract high-precision object edge contours within single frames. 
To further refine these features, we apply the multi-frame weighting strategy for better feature selection. 
To evaluate the impact of the strategy, we run two separate tests under identical conditions— one using the strategy and the other not using it.
Our experimental results, summarized in Tab.~\ref{tab:ablation}, demonstrates that the implementation of the multi-frame weighting strategy leads to an improvement in calibration accuracy, with a $60.30\%$ enhancement in translation accuracy and a $61.26\%$ gain in rotation accuracy.

\subsection{Experiment on our own dataset}
Additionally, we conducted tests on data we recorded ourselves,
including both indoor garages and outdoor campus environments. We compared the results obtained using our method against traditional checkerboard-based calibration methods~\cite{zhou2018automatic}. As shown in Tab.~\ref{tab:ourdata}, our method achieves similar accuracy to the checkerboard-based one without using a checkerboard, which demonstrates the flexibility of our proposed method. The differences ($\Delta$) between the two methods are within acceptable limits, highlighting the comparable performance of our approach.

\begin{table}[tbp]
    \caption{Comparison of accuracy between our method and checkerboard-based calibration method.}
    \centering
    \begin{tabularx}{\linewidth}{>{\centering\arraybackslash}p{0.8cm} *{3}{X} *{3}{X}}
        \toprule
        \multirow{2}[2]{*}{Setting} & \multicolumn{3}{c}{Translation(\si{\m})} & \multicolumn{3}{c}{Rotation(\si{\degree})} \\
        \cmidrule(lr){2-4} \cmidrule(lr){5-7}
         & X & Y & Z & Roll & Pitch & Yaw \\ 
        \midrule
        \cite{zhou2018automatic} & $0.023$ & $-0.453$ & $-0.637$ & $-11.614$ & $-84.964$ & $101.852$ \\
        ours & $0.039$ & $-0.442$ & $-0.619$ & $-11.712$ & $-84.945$ & $101.976$ \\
        $\Delta$ & $0.016$ & $0.011$ & $0.018$ & $0.102$ & $0.019$ & $0.124$ \\
        \bottomrule
    \end{tabularx}
    \label{tab:ourdata}
    \vspace{-1.0em}
\end{table}




%% file: 5_conclusion.tex
\section{CONCLUSION}
\label{sec:conclusion}
Accurate extrinsic calibration between LiDAR and cameras is pivotal for multimodal data fusion tasks. In this paper, we introduce EdgeCalib, a novel online calibration method that utilizes edge features. A significant innovation of our approach lies in the utilization of the SAM visual foundational model, coupled with a multi-frame weighting strategy. These elements collectively establish accurate and robust feature correspondences. Experimental results on both the KITTI public dataset and our own data demonstrate the efficacy and robustness of our method. For future work, we aim to further enhance the algorithm by incorporating lighter and more robust modules.

%% file: root.bbl
\begin{thebibliography}{10}
\providecommand{\url}[1]{#1}
\csname url@samestyle\endcsname
\providecommand{\newblock}{\relax}
\providecommand{\bibinfo}[2]{#2}
\providecommand{\BIBentrySTDinterwordspacing}{\spaceskip=0pt\relax}
\providecommand{\BIBentryALTinterwordstretchfactor}{4}
\providecommand{\BIBentryALTinterwordspacing}{\spaceskip=\fontdimen2\font plus
\BIBentryALTinterwordstretchfactor\fontdimen3\font minus \fontdimen4\font\relax}
\providecommand{\BIBforeignlanguage}[2]{{%
\expandafter\ifx\csname l@#1\endcsname\relax
\typeout{** WARNING: IEEEtran.bst: No hyphenation pattern has been}%
\typeout{** loaded for the language `#1'. Using the pattern for}%
\typeout{** the default language instead.}%
\else
\language=\csname l@#1\endcsname
\fi
#2}}
\providecommand{\BIBdecl}{\relax}
\BIBdecl

\bibitem{zhu2022vpfnet}
H.~Zhu, J.~Deng, Y.~Zhang, J.~Ji, Q.~Mao, H.~Li, and Y.~Zhang, ``Vpfnet: Improving 3d object detection with virtual point based lidar and stereo data fusion,'' \emph{IEEE Transactions on Multimedia}, 2022.

\bibitem{li2022mathsf}
Y.~Li, J.~Deng, Y.~Zhang, J.~Ji, H.~Li, and Y.~Zhang, ``{EZFusion}: A close look at the integration of lidar, millimeter-wave radar, and camera for accurate 3d object detection and tracking,'' \emph{IEEE Robotics and Automation Letters}, vol.~7, no.~4, pp. 11\,182--11\,189, 2022.

\bibitem{graeter2018limo}
J.~Graeter, A.~Wilczynski, and M.~Lauer, ``{LIMO}: Lidar-monocular visual odometry,'' in \emph{Proceedings of 2018 {IEEE}/{RSJ} International Conference on Intelligent Robots and Systems ({IROS})}.\hskip 1em plus 0.5em minus 0.4em\relax {IEEE}, oct 2018.

\bibitem{zhang2004extrinsic}
Q.~Zhang and R.~Pless, ``Extrinsic calibration of a camera and laser range finder (improves camera calibration),'' in \emph{Proceedings of 2004 {IEEE}/{RSJ} International Conference on Intelligent Robots and Systems ({IROS}) ({IEEE} Cat. No.04CH37566)}.\hskip 1em plus 0.5em minus 0.4em\relax {IEEE}, 2004.

\bibitem{levinson2013automatic}
J.~Levinson and S.~Thrun, ``Automatic online calibration of cameras and lasers,'' in \emph{Robotics: Science and Systems {IX}}.\hskip 1em plus 0.5em minus 0.4em\relax Robotics: Science and Systems Foundation, jun 2013.

\bibitem{wang2020soic}
W.~Wang, S.~Nobuhara, R.~Nakamura, and K.~Sakurada, ``Soic: Semantic online initialization and calibration for lidar and camera,'' \emph{arXiv preprint arXiv:2003.04260}, 2020.

\bibitem{kirillov2023segment}
A.~Kirillov, E.~Mintun, N.~Ravi, H.~Mao, C.~Rolland, L.~Gustafson, T.~Xiao, S.~Whitehead, A.~C. Berg, W.-Y. Lo \emph{et~al.}, ``Segment anything,'' \emph{arXiv preprint arXiv:2304.02643}, 2023.

\bibitem{zhang2021line}
X.~Zhang, S.~Zhu, S.~Guo, J.~Li, and H.~Liu, ``Line-based automatic extrinsic calibration of {LiDAR} and camera,'' in \emph{Proceedings of 2021 {IEEE} International Conference on Robotics and Automation ({ICRA})}.\hskip 1em plus 0.5em minus 0.4em\relax {IEEE}, may 2021.

\bibitem{yuan2021pixel}
C.~Yuan, X.~Liu, X.~Hong, and F.~Zhang, ``Pixel-level extrinsic self calibration of high resolution {LiDAR} and camera in targetless environments,'' \emph{{IEEE} Robotics and Automation Letters}, vol.~6, no.~4, pp. 7517--7524, oct 2021.

\bibitem{geiger2012automatic}
A.~Geiger, F.~Moosmann, O.~Car, and B.~Schuster, ``Automatic camera and range sensor calibration using a single shot,'' in \emph{2012 {IEEE} International Conference on Robotics and Automation}.\hskip 1em plus 0.5em minus 0.4em\relax {IEEE}, may 2012.

\bibitem{li2022survey}
X.~Li, Y.~Xiao, B.~Wang, H.~Ren, Y.~Zhang, and J.~Ji, ``Automatic targetless lidar--camera calibration: a survey,'' \emph{Artificial Intelligence Review}, pp. 1--39, 2022.

\bibitem{pandey2012automatic}
G.~Pandey, J.~R. McBride, S.~Savarese, and R.~M. Eustice, ``Automatic targetless extrinsic calibration of a 3d lidar and camera by maximizing mutual information,'' in \emph{Proceedings of Twenty-Sixth AAAI Conference on Artificial Intelligence}, 2012.

\bibitem{pandey2015automatic}
------, ``Automatic extrinsic calibration of vision and lidar by maximizing mutual information,'' \emph{Journal of Field Robotics}, vol.~32, no.~5, pp. 696--722, sep 2014.

\bibitem{wang2012automatic}
R.~Wang, F.~P. Ferrie, and J.~Macfarlane, ``Automatic registration of mobile {LiDAR} and spherical panoramas,'' in \emph{Proceedings of 2012 {IEEE} Computer Society Conference on Computer Vision and Pattern Recognition Workshops}.\hskip 1em plus 0.5em minus 0.4em\relax {IEEE}, jun 2012.

\bibitem{taylor2015motion}
Z.~Taylor and J.~Nieto, ``Motion-based calibration of multimodal sensor arrays,'' in \emph{Proceedings of 2015 {IEEE} International Conference on Robotics and Automation ({ICRA})}.\hskip 1em plus 0.5em minus 0.4em\relax {IEEE}, may 2015.

\bibitem{schneider2017regnet}
N.~Schneider, F.~Piewak, C.~Stiller, and U.~Franke, ``{RegNet}: Multimodal sensor registration using deep neural networks,'' in \emph{Proceedings of 2017 {IEEE} Intelligent Vehicles Symposium ({IV})}.\hskip 1em plus 0.5em minus 0.4em\relax {IEEE}, jun 2017.

\bibitem{iyer2018calibnet}
G.~Iyer, R.~K. Ram, J.~K. Murthy, and K.~M. Krishna, ``{CalibNet}: Geometrically supervised extrinsic calibration using 3d spatial transformer networks,'' in \emph{Proceedings of 2018 {IEEE}/{RSJ} International Conference on Intelligent Robots and Systems ({IROS})}.\hskip 1em plus 0.5em minus 0.4em\relax {IEEE}, oct 2018.

\bibitem{lv2021lccnet}
X.~Lv, B.~Wang, Z.~Dou, D.~Ye, and S.~Wang, ``{LCCNet}: {LiDAR} and camera self-calibration using cost volume network,'' in \emph{Proceedings of 2021 {IEEE}/{CVF} Conference on Computer Vision and Pattern Recognition Workshops ({CVPRW})}.\hskip 1em plus 0.5em minus 0.4em\relax {IEEE}, jun 2021.

\bibitem{liao2023se}
Y.~Liao, J.~Li, S.~Kang, Q.~Li, G.~Zhu, S.~Yuan, Z.~Dong, and B.~Yang, ``Se-calib: Semantic edges based lidar-camera boresight online calibration in urban scenes,'' \emph{IEEE Transactions on Geoscience and Remote Sensing}, 2023.

\bibitem{canny1986computational}
J.~CANNY, ``A computational approach to edge detection,'' pp. 184--203, 1987.

\bibitem{iverson1962programming}
K.~E. Iverson, ``A programming language,'' in \emph{Proceedings of the May 1-3, 1962, spring joint computer conference}, 1962, pp. 345--351.

\bibitem{geiger2013vision}
A.~Geiger, P.~Lenz, C.~Stiller, and R.~Urtasun, ``Vision meets robotics: The {KITTI} dataset,'' \emph{The International Journal of Robotics Research}, vol.~32, no.~11, pp. 1231--1237, aug 2013.

\bibitem{zhou2018automatic}
L.~Zhou, Z.~Li, and M.~Kaess, ``Automatic extrinsic calibration of a camera and a 3d lidar using line and plane correspondences,'' in \emph{Proceedings of 2018 IEEE/RSJ International Conference on Intelligent Robots and Systems (IROS)}.\hskip 1em plus 0.5em minus 0.4em\relax IEEE, 2018, pp. 5562--5569.

\end{thebibliography}
